\documentclass[preprint,12pt]{elsarticle}

\usepackage{amssymb}
\usepackage{amsthm}

\usepackage{url}
\usepackage{multirow}
\usepackage{soul}
\usepackage{color}
\usepackage{tablefootnote}
\usepackage{lscape} 
\usepackage{tabularx}
\usepackage{booktabs}
\usepackage{array}
\usepackage{subcaption}

\journal{King Saud University Computer and Information Sciences}

\begin{document}

\begin{frontmatter}



\title{Accurate  prediction of international trade flows: Leveraging knowledge graphs and their embeddings}


\author[inst1]{Diego Rincon-Yanez}
\author[inst2]{Chahinez Ounoughi}
\author[inst2,inst3]{Bassem Sellami}
\author[inst4]{Tarmo Kalvet},
\author[inst4]{Marek Tiits}
\author[inst1]{Sabrina Senatore}
\author[inst2]{Sadok Ben Yahia}
\affiliation[inst1]{organization={Department of Information and Electrical Engineering and Applied Mathematics, University of Salerno},
            addressline={Via Giovanni Paolo II 132}, 
            city={Fisciano},
            postcode={84084},
            country={Italy}}

\affiliation[inst2]{organization={Department of Software Science, Tallinn University of Technology},
            addressline={Akadeemia tee 15a}, 
            city={Tallinn},
            postcode={12618}, 
            country={Estonia}}
        
\affiliation[inst3]{organization={Faculty of Sciences of Tunis, University of Tunis El Manar},
            addressline={Campus universitaire}, 
            city={Tunis},
            postcode={1092}, 
            country={Tunisia}}

\affiliation[inst4]{organization={Department of Business Administration, Tallinn University of Technology},
            addressline={Ehitajate tee 5}, 
            city={Tallinn},
            postcode={12618}, 
            country={Estonia}}

\begin{abstract}
Knowledge representation (KR) is vital in designing symbolic notations to represent real-world facts and facilitate automated decision-making tasks. Knowledge graphs (KGs) have emerged so far as a popular form of KR, offering a contextual and human-like representation of knowledge. In international economics, KGs have proven valuable in capturing complex interactions between commodities, companies, and countries. By putting the gravity model, which is a common economic framework, into the process of building KGs, important factors that affect trade relationships can be taken into account, making it possible to predict international trade patterns. This paper proposes an approach that leverages Knowledge Graph embeddings for modeling international trade, focusing on link prediction using embeddings. Thus, valuable insights are offered to policymakers, businesses, and economists, enabling them to anticipate the effects of changes in the international trade system. Moreover, the integration of traditional machine learning methods with KG embeddings, such as decision trees and graph neural networks are also explored.
The research findings demonstrate the potential for improving prediction accuracy and provide insights into embedding explainability in knowledge representation. The paper also presents a comprehensive analysis of the influence of embedding methods on other intelligent algorithms.
\end{abstract}

\begin{highlights}
\item A novel way to exploit Knowledge Graphs to model and predict International Trade.
\item A knowledge graph construction method that enhances the prediction accuracy results by inputting the embeddings as features in the destiny models
\item Visualisation and improved prediction of global trade flows on the level of goods offers vital insights for policy makers, businesses, and economists.
\end{highlights}


\end{frontmatter}

\section{Introduction}
\label{sec:introduction}

One primary objective of knowledge representation (KR) is to focus on designing and improving symbolic notations for expressions and real-world facts \cite{Jarke1978, DAVIS201598} for later help from expert systems in automated decision-making tasks. As a result, KR is critical to offering a simple strategy for defining relevant and contextual information within a finite number of facts from a specific domain of interest; these facts are referred to as a knowledge base (KB).

In the past years, Knowledge Graph (KG), as a form of KR, has gained attention because it provides a contextual, natural, and human-like form of representing knowledge in specific domains and common sense. KG is formed in statements called triples on the $T=(h,r,t)$ form, where $h$ (\emph{head}) and 
$t$ (\emph{tail}) represent objects in real life, and $r$, the \emph{relation} is the connection between those entities. 
Internet companies like Google, Wikipedia, and Facebook have found a simple but powerful unified tool in the KG field to describe their multi-structured and multi-dimensional knowledge base, capturing user data to transform it into vast KBs \cite{cimiano17}.

The KG approach is particularly relevant to studying international trade, a significant cornerstone of economic and social development in the globalized economy \cite{cattaneo2010, ponte2019}.
International trade is complex and interconnected, with multiple entities (commodities, companies, and countries) interacting in multiple ways \cite{worldbank2020}. This method helps to understand those complex interactions in a structured and intuitive way. In international economics, the gravity model, a fundamental part of the current method, is widely used to predict trade relations between entities based on factors like size (GDP, population) and distance or other factors \cite{aucamp2023, bergeijk2010, head2014}. By integrating the gravity model into the knowledge base development process, this method captures the key factors that impact trade relationships. It enables the prediction of trade patterns, which is significant for policymakers and economists who need to estimate the potential effects of local changes such as new trade agreements or tariff adjustments on the overall trade scenario.
Applying this method in foresight can yield compelling insights \cite{tiits2023synergies, tiits2013piggybacking}. Moreover, market intelligence can provide companies with a strategic advantage \cite{nasullaev2020technology}, while understanding international trade trends offers valuable direction on product and market prioritization \cite{kalvet2023good, tiits2023goodtrade}.

The method can also be scaled to handle large data sets, making it suitable for analyzing vast and ever-growing data on international trade. Moreover, the KG approach is flexible and can be adapted to various contexts and enriched with additional data types, such as modes of transport or country risk indicators.

Thus, the availability of bilateral trade data and modern machine learning tools provides a unique opportunity to understand and predict international trade flows at the level of individual products to inform decision-making by governments, policymakers, companies, and other researchers.
This study presents a method for modeling international trade by exploiting contextual and relational information using the natural properties of knowledge graphs. The novelty of this work is manifold:
\begin{itemize}
    \item A KGE model is used for link prediction; the gravity model of trade was encapsulated into a Knowledge Graph representation. Moreover, the TransE model was implemented to create gravity model-empowered embeddings.
\item The generated embeddings reveal hidden insights about the trade data: they are used as input for known ML methods such as linear regression, decision trees, or Graph Neural networks (GNNs) to enhance the performance of selected ML models.
\item At the same time, the embedding provides a synthetic high-level interpretation of the trade data:
are projected in multi-dimensional space to reveal the gravity interaction among the data. This projection can be framed as an initial step toward embedding explainability for knowledge representation purposes.
\end{itemize}

The remainder of this paper is structured as follows: Section \ref{sec:materiasl-methods} describes key baseline concepts and data sources exploited in this research. Section \ref{sec:related-work} provides an overview of related works that use gravity models and traditional machine learning approaches for analyzing international trade. Section \ref{sec:gravity-method} presents the gravity-inspired knowledge base construction method. Then, Section \ref{sec:experiments-results} presents experimental results that highlight the power of embedding in improving ML-based performances for decision trees and graph neural networks. Also, an explainable visual validation and a comparative analysis will be provided. Finally, the conclusions and the future work are highlighted in Section \ref{sec:conclusions}.

\section{Materials and Methods}
\label{sec:materiasl-methods}
KGs have emerged as an effective way to integrate disparate data sources and model the underlying relationships for many applications.
The most striking feature of this work is the use of KGs as the basis for predicting international trade flows. For this purpose, the most comprehensive trade datasets are used.

\subsection{Datasets}
This section presents the main international trade flow analysis sources. 
\subsubsection{International Trade Data}
The United Nations Commodity Trade Statistics Database (UN Comtrade) is the single most authoritative global data source on international trade in goods and services. It is the original, most standardized, and comprehensive database on international goods trade. It covers bilateral trade between 250 countries and territories, with more than $5,000$ products classified according to the Harmonized System (HS). It contains vast information, including import, export, and re-export values, at the 6-digit Harmonized System (HS) product code level \cite{un2023}.

The UN Comtrade dataset spans several decades, providing a comprehensive global trade perspective. Each entry in the database is a structured bilateral trade flow record that includes data on the reporting country, the partner country, the product traded, the trade value, and the quantity. The data is sourced from official national entities, thus ensuring a commendable degree of accuracy and reliability.

Nevertheless, the Comtrade dataset is not without its limitations. Missing data constitutes a challenge, primarily because not all nations promptly report their trade data to the UN. Further inconsistencies can arise, for instance, when the export data from a given country does not align with the import data reported by its trade partner. The classification of goods poses another problem, often due to different revisions of the HS nomenclature or discrepancies in the HS codes assigned to the same product by different entities. Valuation differences further compound these problems. These issues are most prevalent in the context of services and weight-related data \cite{chen2022, jiang2022, zhang2022}, and less so in the realms of physical goods and trade values, which are the primary focus of our research.

The Centre d'Etudes Prospectives et d'Informations Internationales (CEPII) has further improved the UN Comtrade database. In their BACI (Base pour l'Analyse du Commerce International) database \cite{baci2023}, a further harmonization process, which addresses discrepancies and inconsistencies, has been carried out \cite{gaulier2010}. As such, the BACI dataset is considered an even more reliable and consistent source of information regarding international trade and is often considered the data source of choice for gravity models.

In the current research, our analytical focus is constrained to the examination of the data from the years 2015 through 2019. This intentional decision stems from two consecutive events' significant global economic disruptions. The first of these was the unprecedented COVID-19 pandemic, which severely disrupted the international economy during the years 2020 and 2021. Subsequent to the pandemic, the Russian military engagement against Ukraine and broader Western countries, beginning in 2022, likewise brought about substantial economic changes. Both the pandemic and the geopolitical conflict have disrupted typical economic patterns, thereby rendering these particular years outliers in any longitudinal analysis. Given this, including the data from these years could lead to skewed projections and compromise the reliability of our future predictions.

\subsubsection{Gravity Model and Data}

The gravity model is a mathematical model widely used in international economics and trade analysis to predict and explain bilateral trade flows between two countries.
It is used to measure the commercial trade synergies between two countries using specific economic indicators such as GDP, GDP per capita, etc. or geopolitical scores such as distance, population, or trade agreements.
More formally, it describes the attraction force of $G_{ij} $ between two entities in space, $e_ i $ and $e_ j \in E = \{e_1, e_2, e_3, \dots, e_n \} $ the set of entities:

\begin{equation}
    G_{ij} = G \cdot \frac{M_i \cdot M_j}{D_{ij}}
    \label{eq:gravityexpression}
\end{equation}
where $ M_i $ and $ M_j $  are the masses of the entities $e_ i $ and $e_ j$ respectively, and $D_{ij}$ is the distance between them, and $ G $ a gravitational constant. 

This model provides a robust empirical relationship with bilateral trade flows and is widely used in many trade datasets.

The CEPII Gravity Dataset \cite{cepii2022}, for instance, is a comprehensive resource extensively employed in gravity model analyses, a principal method for understanding and examining bilateral trade patterns. This dataset encapsulates various variables, facilitating a deeper understanding of international trade dynamics. Geographic variables include bilateral distances and common borders, while cultural variables capture elements like shared languages or historical colonial ties. Further, the dataset incorporates essential macroeconomic indicators, such as GDP, GDP per capita, and population data. The CEPII Gravity Dataset spans a substantial time frame, from 1948 to 2020, and encompasses a wide geographic coverage, with data about 252 countries and territories.

The data within the CEPII Gravity Dataset is collected from diverse sources. While a significant portion of the data is sourced directly from CEPII's databases, other institutional sources like the World Bank's World Development Indicators (WDI), the International Monetary Fund (IMF), and the World Trade Organization (WTO) also contribute to the richness of the dataset. For example, GDP and population data are sourced from the World Bank’s WDI \cite{conte2022}.

The CEPII Gravity Dataset has proven instrumental in various research contexts, elucidating nuanced aspects of international trade. For instance, it has been utilized to investigate the interplay between countries' participation in global value chains and their positions in the international trade network, demonstrating a strong correlation between these elements \cite{yanikkaya2021}. It has also been used to examine the impact of regulatory burdens on international trade \cite{nabeshima2021}, and to argue that the gravity model performs extremely well for describing bilateral trade in final goods and intermediate inputs \cite{greaney2020}.\subsection{Knowledge graphs}

\label{sec:KG}
This section is devoted to an in-depth exploration of the fundamental concepts behind a KG, commonly viewed as a directed heterogeneous multigraph whose node types and relationships have domain-specific semantics. 

KGs allow knowledge to be encoded in a form that can be interpreted by humans and is amenable to automated analysis and inference. KGs are becoming a popular approach for representing diverse types of information in the form of different types of entities connected via different types of relations.

\subsubsection{KG: Basic notions}

Knowledge Graphs (KGs) are a method of representing knowledge at various levels of abstraction and granularity. 
KGs are directed data-based network models whose edges can represent semantic properties; they can be classified into (1) homogeneous KG \cite{kg-book} $ G=(V, E) $, where all the edges in $ E $ are of the same type, i.e., one relation type is defined between the vertices $ V $; while in (2) heterogeneous KG $ G=(V, E, R) $ \cite{Lee2020, kg-book}, where R is a set of edge types or relation labels. 

A Knowledge base (KB) can be represented as a network of atomic assertions known as facts arranged in a statement list, $ KG=\{St_1, St_2, St_3, \dots, St_n \} $. Each statement $St_j$ (j=1 \dots, n) is a triple composed by a  \textit{subject, predicate} and \textit{object}, $ T=<s, p, o>$ (or $<h, r, t> $ , i.e., \textit{head, relation, tail}, respectively) where $s$ and $o \in V$ and  $p \in E$.
 
\subsubsection{Knowledge Graph Embeddings}

\begin{figure}[t]
\centering
\includegraphics[width=0.3\textwidth]{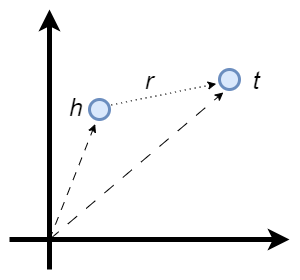}
\caption{Embedding space representation of a single triple $ (h,r,t) $}
\label{Fig:KGE-TransE}
\end{figure}

Knowledge Graph Embeddings (KGEs) are supervised learning models that generate vector representations of the labeled directed multigraph edges; these learn low-dimensional representations based on the entities and relations to predict missing facts \cite{DiPaolo2023}. Given an incomplete knowledge base, one possible task is to predict unknown links. KGE models achieve this through a scoring function $\mathcal{L}$ that assigns a score $ e = \mathcal{L}(h, r, t) \in \mathbb{R} $, indicating whether a triple is true, intending to score all missing triples correctly. The score function $ \mathcal{L}(h, r, t) $ measures the salience of a candidate triple (h, r, t). The optimization goal is usually to assign a higher score to true triples $ (h, r, t) $ than to corrupted false triples $ (h, r' t) $ or $ (h, r’, t') $. Let us remark that a triple one between the head and the tail can be corrupted (denoted by a superscript).
The KGE is commonly used for link prediction; the task focuses on the missing part of a triple against a specific KB that was trained. In the dimensional embedding space, these KGE models are denoted by various score functions \cite{Ristoski2019} that quantify the distance between two entities through the relation type, as shown in Fig \ref{Fig:KGE-TransE}. The KGE models are trained using these score functions, so entities connected by relations are close to one another, and entities without connections are far away.

These embeddings use the same dimensional space to represent entities and relationships. Moreover, all the models use a scoring function to assess the plausibility of a triple. Finally, they minimize an objective function that evaluates the discrepancy between the predicted triples and the actual triples in the knowledge base. These models are relatively simple compared to others based on complex architectures, involving multiple layers of neural networks and additional techniques such as attention mechanisms, graph convolutional networks, and tensor factorization \cite{Rossi2021}.

In more formal terms, the model learns vector embeddings of the entities and relationships for each training set $ S $ of triples $ (h, r, t) $ composed of two entities $ h, t \in E$ (the set of entities) and a relationship $ r \in E$. The embeddings take values in $ \mathbb{R} ^ N $ where $ N $ is the space dimensionality. TransE considers the translation of the vector representations, i.e., the head entity embedding should be close to the tail entity embedding plus relation embedding when the head entity is similar to the tail entity (when $ (h,r,t) $ holds). Otherwise, the head entity should be far away from the tail entity. According to the energy-based framework proposed in \cite{NEURIPS2019_378a063b}, the energy of a triple is equal to $ d(h+l,t) $ for some dissimilarity measure $ d $, which can be either the L1 or the L2-norm.

The following section introduces  a review of related work concerning the analysis of international flow data sets.

\section{Related Work}
\label{sec:related-work}

Over the years, trade analysis has gained significant attention due to its essential role in economic growth and development. Researchers have conducted numerous studies to understand the factors influencing trade flows between countries and regions. The related data science research can be broadly categorized into gravity-based and alternative machine-learning-based approaches. The gravity approach utilizes a set of fundamental factors, such as the size of the economies, the distance between trading partners, and cultural similarities, to model trade relationships. On the other hand, alternative machine learning-based approaches use data-driven methods to learn the patterns and relationships within the trade data. Both approaches have strengths and limitations, and researchers continue exploring their potential to improve trade analysis and decision-making. This section reviews some of the significant contributions made in these two categories of trade analysis.

\subsection{Gravity model-based}

Several studies have used the gravity model to analyze trade relations between countries and regions. Worth mentioning, Liu et al. \cite{liugravity} empirically analyzed Chinese apparel export flows using annual UN Comtrade data from 2000–2020. They also used gravity data for the GDPs and distances and evaluated the GDP coefficient and elasticity. Korepanov et al. \cite{korepanov2023statistical} used the annual UN Comtrade database from 1994–2021 to conduct statistical analysis between Latvia and Ukraine. Recently, Mafakheri et al. \cite{mafakheri2023predicting} developed a graphical-based link prediction model to investigate the relations between countries and products for importing and exporting petrochemicals using Comtrade annual data from 2017–2019. They used scoring methods such as CN, AA, JC, RA, PA, Katz, CNGF, and KatzGF. Xu et al. \cite{xu2023reconstruction} compared seven network reconstruction methods for capturing the linkages of the international energy trade network using 15 HS codes of energy trade data extracted from UN Comtrade 2019. The best method across all 15 energy trade networks regarding link-based similarity measures was not specified, but fitness models tended to perform best at reconstructing the weight matrix.
Hansen et al. \cite{hansen2023covid} investigated the COVID-19 pandemic's effect on Chinese trade relations using COVID data and monthly imports and exports from China to all available trading partners in 2020. They used regression, gravity, and panel-data regression estimators to predict Chinese trade flows and differentiate between observed and residual trade. The study found that stricter lockdown measures in export destinations negatively impacted Chinese exports and reduced the difference between observed and predicted trade flows, indicating the potential for higher trade during pandemic periods. Another study by Davidescu et al. \cite{davidescu2022empirical} analyzed Romania's trade performance and found that its exports are influenced by EU demand and imports from China and the rest of the world. They used gravity model formulas with the log between 2020 and 2021 and assessed Romania's ability to recover from the COVID-19 pandemic using simulation forecasting scenarios. Khan et al. \cite{khan2019prediction} proposed using modern portfolio theory to predict optimal export commodities to maximize profit and minimize risk. They used the HS [6-digit code] dataset of all the commodities from 2003 to 2016 and compared portfolio optimization with traditional methods. They also used a quantitative factor for risk involving product complexity and the gravity trade model. The study reported improved trade prediction accuracy using portfolio optimization and classified optimal products for export investment based on the risk level. They used models such as the Markowitz mathematical framework, the Black-Litterman model, and the trade gravity model and used the mean squared error (MSE) for evaluation.

\subsection{Alternative machine learning-based}

Gopinath et al. \cite{gopinath2020machine} used supervised and unsupervised machine learning techniques to predict agricultural trade patterns for seven significant commodities. The authors demonstrated that these models outperformed traditional approaches and provided better long-term predictions. Panford-Quainoo et al. \cite{panford2020bilateral} proposed a framework for predicting bilateral trade partners using graph representation learning. The authors achieved high accuracy for node classification and link prediction tasks by comparing different graph neural network models, including GCN, ChebNet, GAT, AGNN, linear, logistic regression, GAE, and VGAE, from 111 different countries. Hu et al. \cite{hu2022spatiotemporal} identified a new form of inconsistency in the UN Comtrade database called "statistical imbalance." The authors proposed a prescreening method to ensure the validity of research results, particularly in commodity categories like fossil fuels, pharmaceuticals, and machinery. Guo et al. \cite{guo2022chinese} introduced a new gravity model based on a back propagation neural network to predict the export potential of photovoltaic products from China. The model was trained on data from multiple sources, including UN Comtrade, the CEPII-BACI database, IRENA, and WDI. Li et al. \cite{li2023rmultinet} developed the R package rMultiNet. The package uses tensor decomposition techniques and visualization tools to analyze a mixture of multilayer networks. They use the 2019 trade data from UN Comtrade as an experiment use case. Xie et al. \cite{xie2023interpretable} proposed an oil trade decision-making model based on explainable machine learning, game theory, and utility theory. The model considers economies' benefit and cost endowments in international oil trades.

\subsection{Scrutinizing takeaways}
Table \ref{tab:comparison} presents a comparative study of various research papers published on gravity and machine learning models used in international trade analysis.

\begin{landscape}
\begin{table}[ht]
\centering
\caption{Comparative Study of Gravity and Machine-Learning based Trade Analysis.}
\label{tab:comparison}
\scalebox{0.75}{
\begin{tabular}{lcm{3cm}lcm{5cm}m{2.5cm}m{7cm}}
\toprule
\textbf{Category} &\textbf{Pub} & \textbf{Data source} &\textbf{Period}&\textbf{Frequency} & \textbf{HS Codes} & \textbf{Trade Flow} & \textbf{Methodology} \\
\midrule
     \multirow{6}{*}{Gravity-based} & \cite{korepanov2023statistical} &  UN Comtrade & 1994-2021 & Annual & HS codes between Latvia and Ukraine & Bilateral & Statistical analysis \\
    \cline{2-8}
    & \cite{xu2023reconstruction} & UN Comtrade & 2019 & Annual & 15 energy HS codes &  Bilateral & Graph link-based similarity measures\\
    \cline{2-8}
    & \cite{hansen2023covid} & Covid data + UN Comtrade   & 2019-2020 &  Monthly & HS codes between China and all partners & Bilateral & Panel-data regression and Gravity model \\
    \cline{2-8}
    & \cite{mafakheri2023predicting} &  UN Comtrade & 2017-2019 & Annual & 63 Petrochemical HS codes & Bilateral & Graph-based link prediction \\
    \cline{2-8}
    & \cite{liugravity} &  UN Comtrade & 2000-2020 & Annual & HS61 and HS62 between China and 198 partners & Exports & Gravity model, GDP Coefficient and Elasticity \\
    \cline{2-8}
    & \cite{davidescu2022empirical} & UN Comtrade & 2020-2021 & Annual & HS codes between Romania-EU and Romania-China & Bilateral & Panel data gravity model \\
    \cline{2-8}
    & \cite{khan2019prediction} & UN Comtrade & 2003-2016& Annual & 23 textile HS codes between Pakistan and all partners & Exports & Markowitz historical model, CAPM and Black-Litterman model \\
    \bottomrule
    \multirow{5}{*}{Alternative ML-based} & \cite{li2023rmultinet} & UN Comtrade  & 2019 & Annual & 97 HS codes & Exports & Graph embeddings \\
    \cline{2-8}
    & \cite{xie2023interpretable} &UN Comtrade  & 1990-2019 & Annual & HS270900 & Bilateral & Machine learning-based model \\
    \cline{2-8}
   & \cite{guo2022chinese}&UN Comtrade, CEPII-BACI & 2000-2021& Annual & 17 PV HS codes between China and all  partners & Exports &  Back propagation neural network\\
   & & IRENA\tablefootnote{International Renewable Energy Agency}, WDI\tablefootnote{World Development Index} & &  & &  &  \\
    \cline{2-8}
    & \cite{hu2022spatiotemporal} & UN Comtrade & 1996-2016& Annual & 15 countries trades & Bilateral  & Co-clustering algorithm \\
    \cline{2-8}
    & \cite{gopinath2020machine} &UN Comtrade, TRAINS\tablefootnote{UNCTAD Trade Analysis Information System}, & 1962-2020& Annual & 8 agricultural HS codes & Bilateral & LightGBM, MLP, Random forest, Extra tree regression, XGBoost \\
    \cline{2-8}
    & \cite{panford2020bilateral} & UN Comtrade & - & Annual  & 170 countries trades & Bilateral & GCN, ChebNet, GAT, AGNN, Linear and Logistic Regression, GAE, VGAE \\
    \bottomrule
\end{tabular}
}
\end{table}
\end{landscape}
 
A close examination of Table \ref{tab:comparison} allows us to identify the following key items of concern:
\begin{enumerate}
\item More than $99\%$ of approaches utilizing the UN Comtrade data employed a yearly coarse granularity,which limits the availability of instances for machine learning or deep learning models. Consequently, achieving the desired high accuracy becomes challenging. 
 \item The reviewed approaches generally exhibit a narrow scope in terms of the range of commodities and the breadth of HS codes considered.
 \item The scope of the studies themselves is also quite limited. For instance, \cite{korepanov2023statistical} focused solely on bilateral trade flows between Latvia and Ukraine. Other works concentrate only on one-to-many multilateral trade flows, e.g., \cite{liugravity}: China versus 198 partners, \cite{davidescu2022empirical}: Romania versus the European countries.
  \item With the exception of [35] and [32], none of the approaches explored enriching the UN Comtrade data with additional sources. [35] used a deep-learning model, while [32] mainly explored applying classical machine-learning techniques.
\end{enumerate}

To address these shortcomings, a new approach is introduced that exploits the concept of knowledge graph trends to build accurate models for predicting international trade flows. The salient facts of this approach are as follows.
\begin{itemize}
 \item \textbf{Genericity}:  Our approach considers HS-6-digit codes, encompassing more than 5,000 commodities, and multilateral many-to-many international trade flows. As a result, the model's outcomes can be applied by any country.
  \item \textbf{Fine Granularity}: We employ a monthly granularity that offers a detailed perspective over a long time span from 2016 to 2019.
 \item \textbf{Geniuous integration}: We integrate the monthly fine-granularity UN Comtrade data with the yearly coarse-granularity BACI data. To achieve this, a knowledge graph is built and then the associated embedding. This embedding is then incorporated into the UN Comtrade data as an additional feature.
 \item \textbf{Carried out extensive experiments}: Our research involves comprehensive experiments that provide empirical evidence for the advantages of integrating both data sources.

\end{itemize}

\section{Gravity Enhanced KBC Model}
\label{sec:gravity-method}

From a general perspective, the Knowledge Base Construction (KBC) task leverages available data to extract spatial relations among them and discover relevant insights among data features; the goal is to achieve a KR model that encompasses real-world characteristics. 

\begin{figure}[ht]
    \centering
    \includegraphics[width=0.95\textwidth]{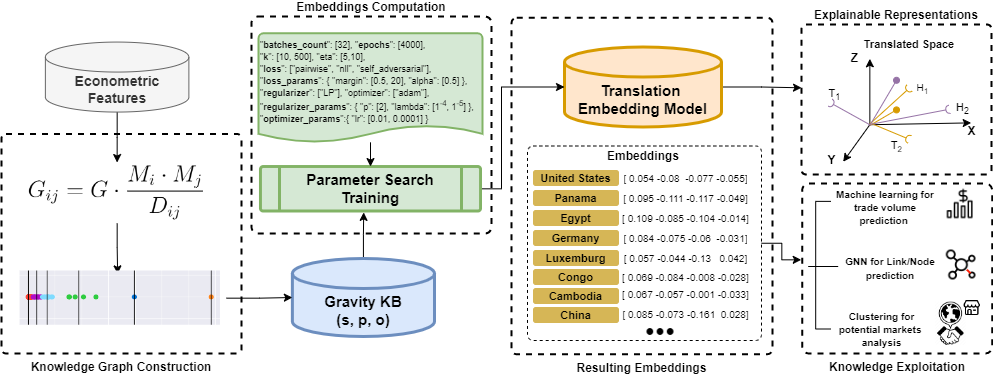}
    \caption{Gravity-based Knowledge Base Construction method presents the data flow. The resulting embeddings can be exported to feed additional IA methods such as Decision Trees and GNN}
    \label{fig:KBC-oveview}
\end{figure}

In Fig. \ref{fig:KBC-oveview}, a logical overview describes the proposed gravity-based KBC model across the interactions among the main components designed to accomplish specific tasks.  

The model takes as input relevant economic Key Performance Indicators (KPI), called {econometric features}. The \emph{Knowledge Graph Construction} component is in charge of processing these features to evaluate the gravity score (using Eq. 1) among entity pairs $s$ and $o$  and generate a triple-based representation $<s, p, o>$, according to the evaluated gravity score $p$.

Then, to categorize the types of gravitational forces among the entities, a clustering method was performed to obtain score partitioning to categorize the different gravitational forces that emerged in the data. The partitioning allows us to generate a linguistic label for the predicate $p$ in our triples $<s, p, o>$.
So computed triples compound the initial Knowledge Graph (KG), according to the definition provided in Section \ref{sec:KG}, or Gravity Knowledge Base (\emph{Gravity KB}), as shown in Fig. \ref{fig:KBC-oveview}.

Once the KG is built,  the \emph{Embeddings Computation} component accomplishes a parameter search optimization task that allows the  \emph{Resulting Embeddings} component to generate, by translational space-based methods,  a  low-dimensional vector entity representation; these vectors are projected in the final translated space for an explainability purpose (\emph{Explainable Representation}); each particular component/task is described in depth in the following subsections. The resulting embedding is used to feed the traditional machine-learning models (\emph{Knowledge Exploitation} component) to leverage the intrinsic information revealed by the embedding and enhance the prediction capability of the models. 

\subsection{Knowledge Graph Construction}

Considering the notions about \emph{triple} and \emph{KG} provided in Section \ref{sec:KG}, our model maps the triple $ T=<s,p,o> $ establishing a direct relationship between two entities $ s \rightarrow e_i $ and $ o \rightarrow e_j $, obtaining the following final triple expression $ T=<e_i, G_{ij}, e_j>$, where $G_{ij}$ is the attraction force, as given in Eq. \ref{eq:gravityexpression}.

\begin{figure}[ht]
    \centering
    \includegraphics[width=0.5\textwidth]{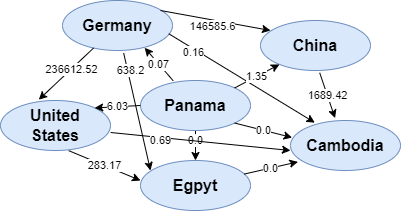}
    \caption{Example of a graph based on trade flows: countries are entities, and relationships are gravity scores}
    \label{fig:KG-raw}
\end{figure}

Figure \ref{fig:KG-raw} presents a trade flow-based graph with countries as nodes and gravity scores as edges between node pairs.

As mentioned, the edge or relationship between nodes in the graph represents a gravity score, that is, a numerical value. An automatic clustering-based process allows score partitioning by simply mapping value ranges into categories or classes to avoid manually classifying these gravity forces among countries.

\paragraph{Clustering for the category generation:}  
The numerical values of the gravity score could be very tricky to model in the graph-based representation, where the relation between subject and object is a label and describes a kind of correlation among entities. Defining as many classes (or categories) as the score values could not be effective for the KGE process.
To deal with this issue and avoid bias in the categorization, a clustering method is applied to the gravity scores, yielding a partitioning whose clusters can be mapped to some categories.

To this purpose, in the range of clustering families \cite{Jain1999}, five clustering methods were considered were carried out. In Table \ref{tab:Clustering}, the selected algorithm and the families to which they belong are mentioned.

\begin{table}[ht] 
\centering
\caption{Implemented Clustering methods with family type and suited designed dataset constraints.}
\label{tab:Clustering}
    \begin{tabular*}{0.92\textwidth}{lcr}
    \toprule
        \textbf{Method} & \textbf{Family} & \textbf{Suited Data} \\
    \midrule
        K-means & Centroid-Based & Large even datasets \\
        Dendogram & Hierarchical & Small and Medium datasets  \\
        DBSCAN & Density-based & Large with arbitrary shapes  \\
        Gaussian Mixture & Model-based & Normally Distributed Data \\
        Mean Shift & Kernel-based & No particular distribution \\
    \bottomrule
    \end{tabular*}
\end{table}

The algorithms cover various requirements regarding data distribution, complexity, and dataset size.
Being the simplest to understand and implement, K-means commonly assumes spherical clusters of equal size; this method is designed for evenly distributed large datasets. On the other hand, Hierarchical clustering, such as dendrogram, does not have any assumptions on shapes and sizes and, therefore, performs well on unevenly distributed clusters; besides, it is computationally expensive, making it optimal to work with small to medium-sized datasets. DBSCAN is a density-based technique that finds clusters of any shape and size but requires tuning for parameters set, in contrast with Dendogram, which performs well with large datasets in arbitrary cluster shapes and sizes. Finally, Mean-Shift follows a more unattended approach with no cluster assumption regarding shape and size, automatically deciding the number of clusters; it is computationally expensive and performs well with small to medium-sized datasets with non-linear cluster shapes.

\subsection{Embeddings Training}

Knowledge Graph embeddings (KGE) are typically used for downstream tasks \cite{Rincon-Yanez2023} like link prediction and entity recognition. However, recent research on NeuroSymbolic IA trends \cite{Besold2017, Rincon-Yanez2022, Breit2023} shows the embeddings as an additional step into the KR approaches and not a downstream task.

Energy-based models are founded on the hypothesis of benefiting from a given function $ E(x) $ by assigning low energy values to some inputs and high energy values to others \cite{LeCun2006ATO}. Similarly, the translational space embedding model, also known as TransE \cite{Bordes2013}, is created with the perception that the head entity $ h$ is close to the tail entity $t$ embeddings through the relationship $ l $. TransE exploits the hierarchical relationship concept, meaning the distance between two connected entities through a relationship is small since they share similar attributes. The loss function for the TransE model \cite{Bordes2013} is described as follows:

\begin{equation}
    \mathcal{L} = \sum_{(h,l,t) \in S} \sum_{(h',l',t') \in S'_{(h,l,t)}} [ \gamma + d(h+l,t) - d(h'+l,t')]_+  
    \label{eq:transe-loss}
\end{equation}

\begin{equation}
    S'_{(h,l,t)} = \{(h',l,t)|h \in \mathcal{E} \} \cup \{(h,l,t')|t' \in  \mathcal{E}\} .
    \label{eq:transe-negative}
\end{equation}

Where $S$ is the set of positive triples in the training set, $S'$ is the negative triples, defined in Eq. \ref{eq:transe-negative} shadowing \emph{head} or \emph{tail} and  $ d(h+l,t)$ is the distance measure between the entities.

TransE provides a simple but powerful framework for capturing semantic associations between entities and relationships in a KG, in contrast to more advanced models such as DistMult \cite{Yang2015} and ComplEx \cite{trouillon16}, which use high-dimensional spaces; HolE \cite{Nickel16} which uses circular correlation in tensor spaces; RotatE \cite{sun2018rotate} employs complex-valued rotation matrices, or the convolutional approach of ConvE \cite{ConvE}.

TransE provides a simple but powerful framework for capturing semantic associations between entities and relationships in a KG, in contrast to more advanced models such as DistMult \cite{Yang2015} and ComplEx \cite{trouillon16}, which use high-dimensional spaces; HolE \cite{Nickel16} which uses circular correlation in tensor spaces; RotatE \cite{sun2018rotate} employs complex-valued rotation matrices, or the convolutional approach of ConvE \cite{ConvE}.

\section{Experiments and Results}
\label{sec:experiments-results}

During the experimental setup of the machine learning models, we performed a comparative analysis to investigate the impact of utilizing the embedding components of both origin and destination countries, which gravity-based KBC created. The models were trained on the UN Comtrade and BACII datasets, consisting of various features such as year, month, GDP, harmonic distance, and commodity code, to predict the trade volume between the two countries. Table \ref{tab:params} provides a comprehensive overview of the dataset used in these experiments.

\subsection{Leveraging gravity embedding features for empowering Decision Tree}
\label{sec:embedding-features}
The candidate model for our first experience was the Decision Tree (DT) model. The model creates a tree structure that captures the relationships between the input features and the target variable. Thanks to the generation of the tree structure, it is easily understood, allowing users to gain insights into the decision-making process.

\begin{figure}[ht]
    \centering
    \includegraphics[width=0.80\textwidth]{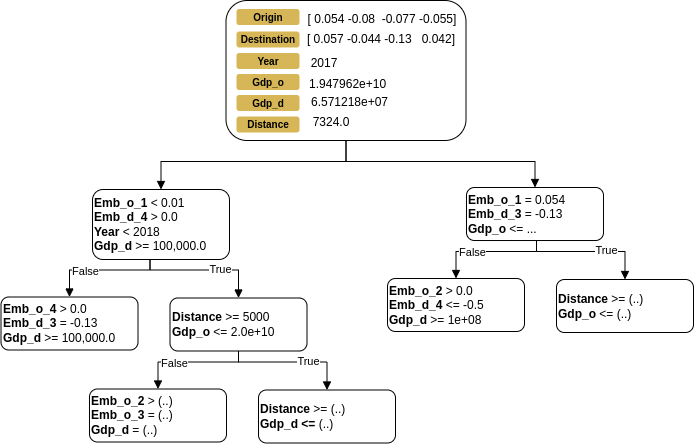}
    \caption{Decision Tree model using the KBC embedding results for trade volume prediction between USA and LUX example.}
    \label{fig:DT-example}
\end{figure}

Fig. \ref{fig:DT-example} shows an example decision tree model for predicting the trade volume between the United States and Luxembourg.

The model uses the embedding components of origin and destination countries created by gravity-based KBC and several other features, including the year, GDP, distance, and commodity code, to predict the trade volume. The decision tree is structured with a series of nodes that split the data based on the values of these features. At each node, the model chooses the feature that provides the most information gain for the prediction task. The tree's leaf nodes contain the predicted trade volume for each combination of feature values. This decision tree model can predict the trade volume between the United States and Luxembourg for any combination of feature values by traversing the tree from the root to a leaf node and using the corresponding predicted value.

\begin{table}[htbp]
\centering
\caption{Experimental setup parameters.}
\begin{tabular}{l|l}
\hline
\textbf{Parameter} & \textbf{Value} \\
\hline
Dataset size & $367,315,312$\\
Reporter codes & 140 \\
Partner codes & 247 \\
Commodity codes & 5,204 \\
Train/Test & $80\%/20\%$  \\
Decision tree & Max depth: 50\\ 
Trade flow & Exports \\
\hline
\end{tabular}
\label{tab:params}
\end{table}

\begin{table}[ht]
    \centering
    
    \caption{Decision Tree performance evaluation.}
    \scalebox{0.95}{
    \begin{tabular}{m{4.2cm}cccc}
        \toprule
        \textbf{Metric} &  \textbf{MAE} & \textbf{MAPE} & \textbf{MPE} & \textbf{R-square}\\\midrule
         Basic features &0.6745 &633,799.3726& -633,784.5952&  0.5276\\ 
         Basic features with log & 0.3661 &\textbf{18,981.6164} & \textbf{-18,951.7242}& 0.5143 \\ 
         Embedding features  &   0.3554 & 580,450.3465 & -580,438.4712 & 0.5564\\
         Embedding features with log  &  \textbf{0.1229} & 657,425.2525 & -546,900.2780 & \textbf{0.6826}\\
        \bottomrule
    \end{tabular}}
    \label{tab:comparativeStudy}
\end{table}

Table \ref{tab:comparativeStudy} presents the performance evaluation of the decision tree model with different feature variations based on various metrics. The first variation experiment used basic features, including trade value, country codes, year, month, commodity code, harmonic distance, and GDP. A logarithm function was applied to the log variation's trade value, harmonic distance, and GDP. Both countries' embedding vectors were used alongside the commodity code for training to predict the trade value of embedding features and embedding features with the log. The evaluation used four metrics: MAE, MAPE, MPE, and R-square. The results indicate that the embedding features of the log model outperformed all other models in terms of MAE and R-square, which suggests that this model has the smallest prediction errors and the highest goodness of fit. On the other hand, the basic features model had the highest MAPE and MPE values, indicating that this model's predictions deviated most significantly from the actual values.

Overall, the evaluation suggests that the embedding features with the log model are the best-performing model, indicating that Gravity-based KBC embedding features and log transformation can significantly improve the model's performance in predicting the target variable. Therefore, it can be concluded that incorporating embedding features and log transformations can enhance different machine learning models' accuracy in predicting trade value.
\begin{figure}[ht!]
     \centering
     \begin{subfigure}{.5\textwidth}
         \centering
         \includegraphics[width=1\textwidth]{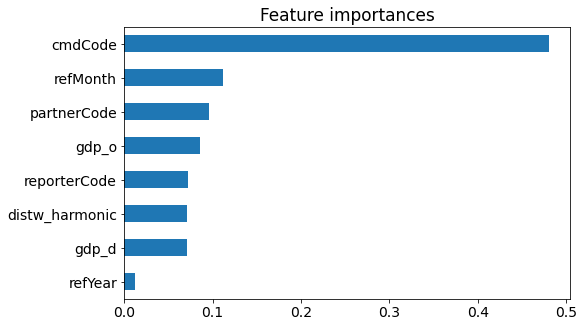}
         \caption{Basic features}
     \end{subfigure}\hfill
     \begin{subfigure}{.5\textwidth}
         \centering
         \includegraphics[width=1\textwidth]{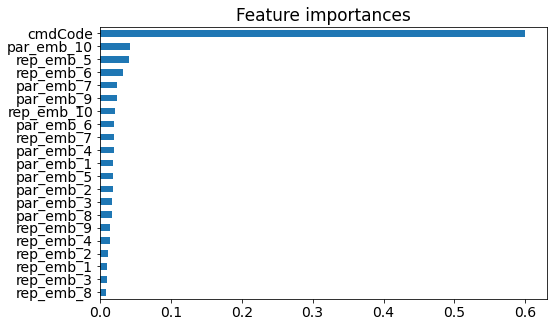}
         \caption{Embeddings features}
     \end{subfigure}
     \caption{Decision Tree feature importance.}
     \label{fig:featureimportance}
 \end{figure}

Figure \ref{fig:featureimportance} shows the feature importance results of the two decision tree models with log transformation trained with basic features and embedding features, respectively. The DT basic features model shows that the commodity code is the most critical feature with an importance value of $0.48$, followed by the partner code with 0.10, and the reference month with $0.11$. However, other features, such as the reporter code, harmonic distance, and GDP, have relatively low importance values compared to the commodity code. In contrast, the DT with embeddings model indicates that the commodity code is still the most crucial feature, with an importance value of 0.60. Embedding features, however, have a higher importance value than basic features, ranging from $0.01$ to $0.04$ for the ten embedding features. This suggests that the embedding features are crucial to improving the model's performance. Additionally, the embedding features for the partner and reporter codes have similar importance values, with the top features having similar importance values. These feature importance results suggest that incorporating embedded features improves the model's performance in predicting trade value.
The embedding features capture more detailed information about the partner and reporter countries, leading to better prediction accuracy.
However, the commodity code remains the most critical feature in predicting the trade value, regardless of the type of features used.

\subsection{Graph Neural Networks (GNNs) and the gravity embedding features}

\begin{figure}[ht]
    \centering
    \includegraphics[width=0.80\textwidth]{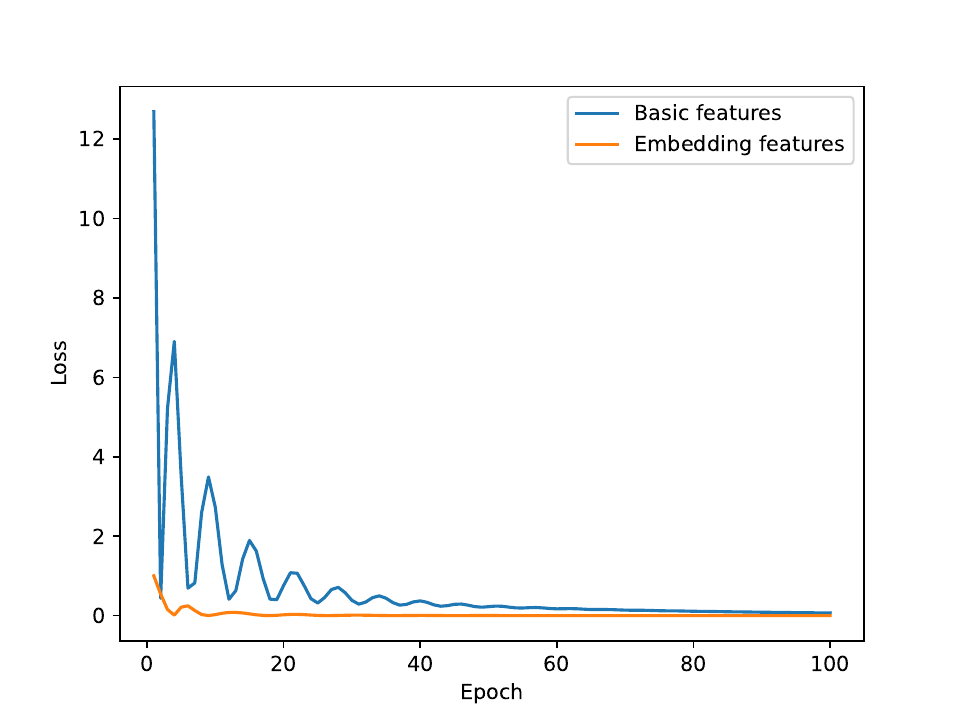}
\caption{MSE Loss function for GNN-based Link Prediction}
\label{fig:GNN-training}
\end{figure}

Figure~\ref{fig:GNN-training} represents the results obtained when gauging the MSE loss function for the GNN-based link prediction tasks while training the original gravity dataset and another training using our embedding features.
The MSE values reported in this figure indicate that the GNN regression model based on embedding features has quickly decreased to $0$; therefore, it has significantly outperformed the GNN model based on the basic features.

\begin{figure}[ht]
    \centering
    \begin{subfigure}{0.49\textwidth}
        \includegraphics[width=\textwidth]{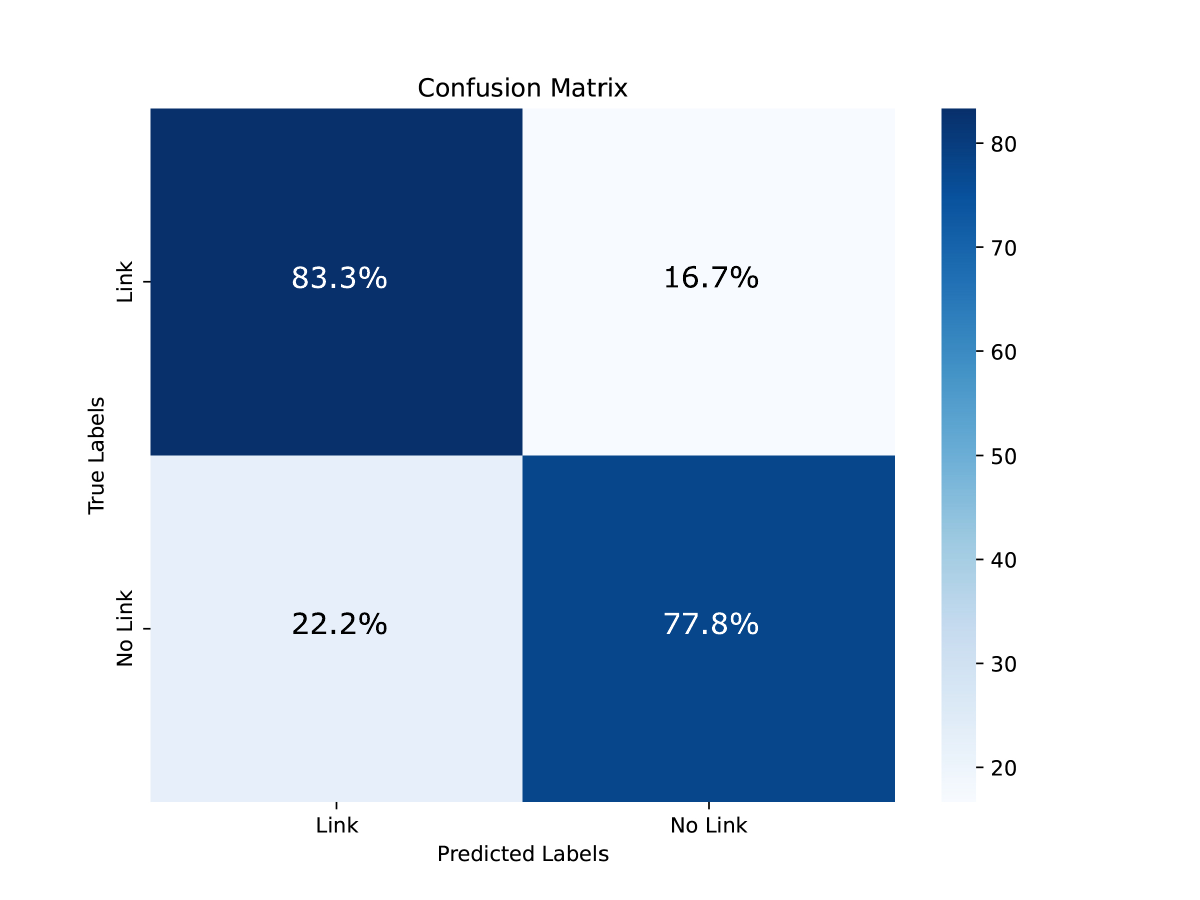}
        \caption{with basic features}
        \label{fig:Mtrx_bc}
    \end{subfigure}
    \begin{subfigure}{0.49\textwidth}
        \includegraphics[width=\textwidth]{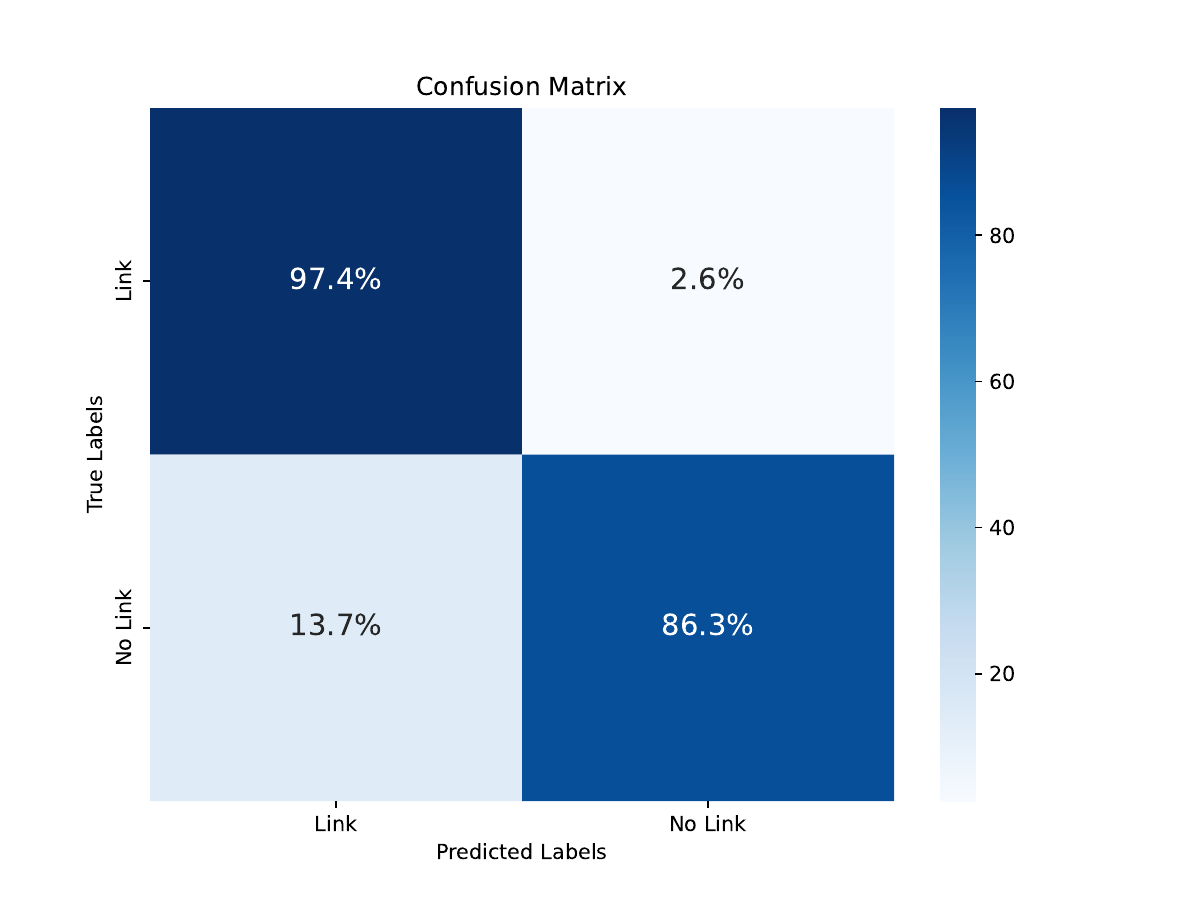}
        \caption{with embedding features}
        \label{fig:Mtrx_emb}
    \end{subfigure}
\caption{Confusion Matrix for GNN-based Link Prediction.}
\label{fig:Mtrx}
\end{figure}

On the other hand, Figure~\ref{fig:Mtrx_bc} represents the confusion matrix of the GNN model with basic features that provide interesting insights. It achieved a true positive rate (TP) of $83.3$\%, meaning it correctly identified a significant portion of true links. However, the false positive rate (FP) was $16.7$\%, indicating a tendency to identify non-existent links as positive mistakenly. Moreover, the false negative rate (FN) was $22.2$\%, implying that the model failed to recognize some true links. On the positive side, the true negative rate (TN) reached $77.8$\%, denoting a reasonably high accuracy in identifying non-links.

In contrast, Figure~\ref{fig:Mtrx_emb}, which represents the confusion matrix of the GNN model with embedding features, demonstrated substantial improvements. It achieved a TP rate of $97.4$\%, indicating its ability to predict true links accurately. The FP rate dropped to $2.6$\%, reflecting a reduced tendency to identify non-existent links incorrectly. However, the FN rate remained at $13.7$\%, suggesting that some true links were still missed. Nevertheless, the TN rate rose to $86.3$\%, highlighting the model's improved accuracy in identifying non-links. It is also shown in table~\ref{tab:cmp_gnn}, which represents the accuracy measured for two models, indicates that the GNN model based on our embedding features has a much higher accuracy value, with a mean of almost $0.9185$ against the other model.

Comparing the two models, it is clear that the GNN model based on embedding features outperformed the one based on basic features. The embedding-based model's higher TP and TN rates signify improved accuracy in predicting both positive and negative links. Although the FN rate is still relatively higher than the TP rate, it is significantly lower than that of the basic features model.

\begin{table}[ht]
    \centering   
    \caption{Accuracy for link prediction task}
    \scalebox{0.95}{
    \begin{tabular}{m{4.2cm}cc}
        \toprule
        \textbf{} &  \textbf{Accuracy} \\\midrule
         Basic features &0.8055\\ 
         Embedding features  & 0.9185\\ 
        \bottomrule
    \end{tabular}}
    \label{tab:cmp_gnn}
\end{table}

Overall, the fact that the GNN regression model, based on the embedding features, outperforms the original dataset regarding predictive accuracy suggests that a model is a promising approach for the given task.

\subsection{Towards an explainability approach using KGE}
\label{sec:explainability-embedding}

\begin{figure}[htb]
    \centering
    \begin{subfigure}{0.52\textwidth}
        \includegraphics[width=\textwidth]{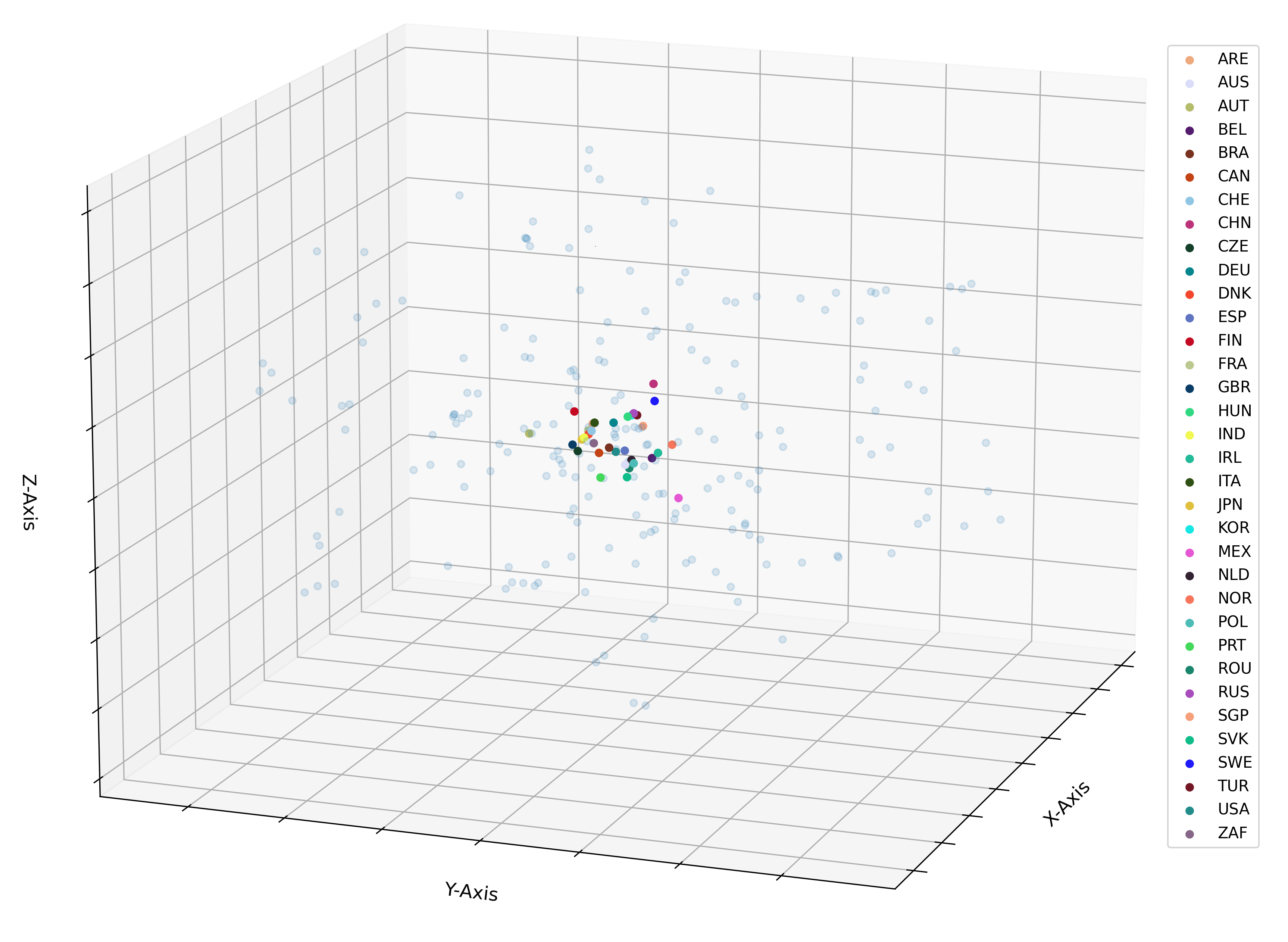}
        \caption{Left View}
        \label{fig:Left}
    \end{subfigure}
    \begin{subfigure}{0.47\textwidth}
        \includegraphics[width=\textwidth]{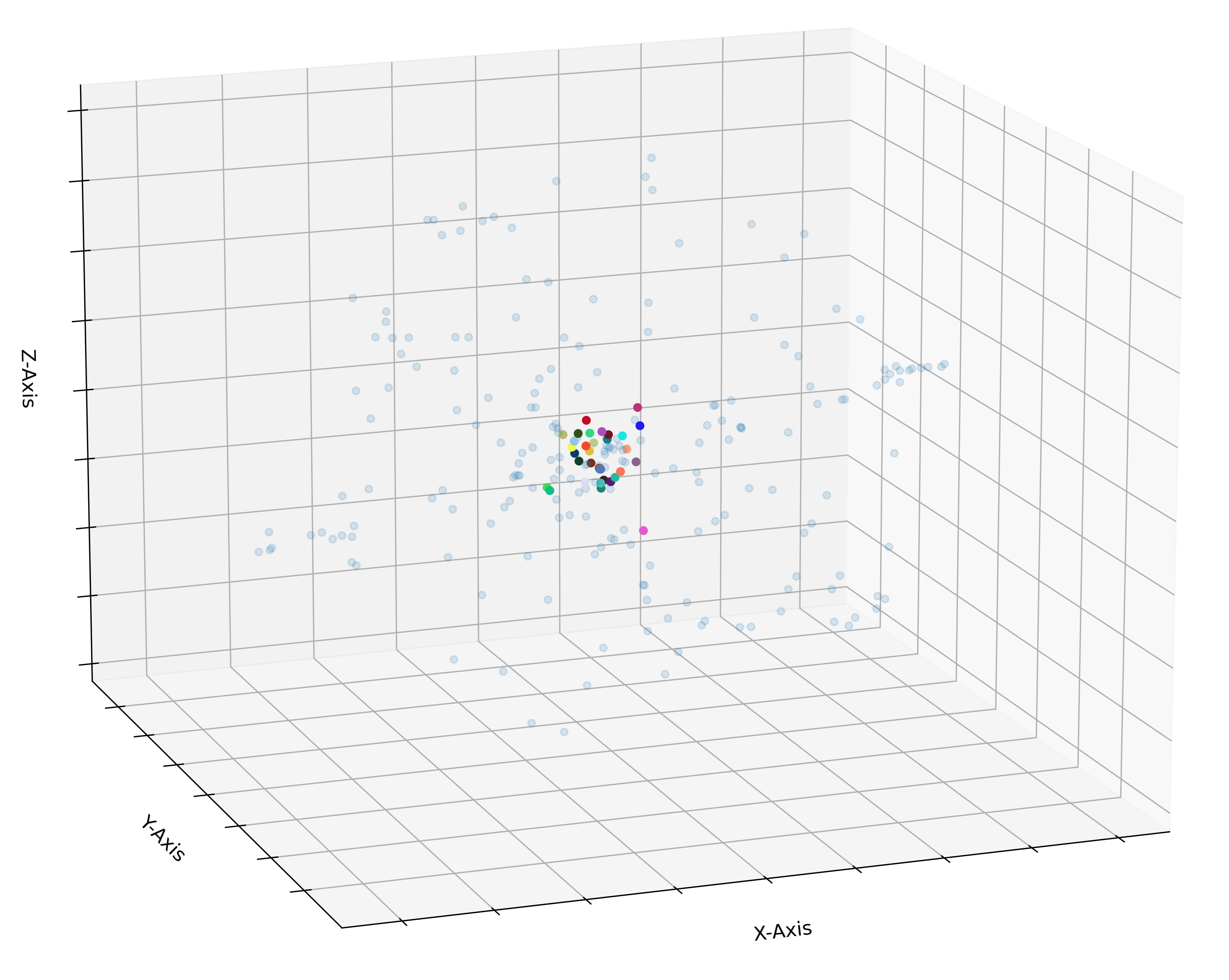}
        \caption{Rigth View}
        \label{fig:Rigth}
    \end{subfigure}
\caption{Isometric view of translational embeddings model projection in a 3D space}
\label{fig:KGE-Scatter3D}
\end{figure}

In large-scale and real-world decisions, understanding how an IA model makes decisions is a mandatory requirement; this is why explainability approaches are becoming more popular daily. For this scenario, the space representations displayed in Fig. \ref{fig:KGE-Scatter3D} provide a comprehensive view of the previous gravity scores and the relations between the nations.

Two three-dimensional plots depicting the same embeddings are presented from two distinct perspectives, enhancing depth perception and enabling a better understanding of the varying "distances" between entities.

In Figure \ref{fig:KGE-Scatter3D}, the international trade closeness of Germany (DEU) for 2019 is portrayed. The primary trade partners include, for example, the United States (USA), France (FRA), China (CHN), the Netherlands (NLD), the United Kingdom (GBR), Italy (ITA), Poland (POL), and Austria (AUT). This illustrative figure also provides a portrayal of the interconnectedness amongst these principal trade partners as well as with other economies globally. This three-dimensional representation facilitates an in-depth comprehension of the global trade ecosystem from Germany's perspective. It also unveils the primary clusters of countries with which Germany does most of its trade transactions and how those countries are connected to other countries around the globe through trade.

This way, our graphical representation visually displays clusters or groups of countries participating in substantial trade relationships. The visualization provides a platform to comprehend the formation of global economic blocs based on trade interactions. It allows for straightforward and accessible recognition of key players and hubs within the trade network, thereby enabling the identification of potent regional economies and influential trading nations.

Moreover, a comparison of these maps with analogous graphical representations from previous years enhances our understanding of the shifting landscape of international trade. Such a temporal analysis sheds light on the dynamism inherent in trade relations, tracing the evolution of trading blocs and transformations in trade patterns over the years.

Additionally, using our predictive modeling can add a prospective dimension to this analysis as we can anticipate future trade trends and shifts in global trade configurations. The subsequent comparison of the projected trade network with the existing network would provide an intriguing exploration of international trade's potential trajectory, contributing significantly to our understanding of global economic development and growth patterns.

Moreover, an added value of our work lies in the capability to create more product-centric representations, be it at the level of the detailed six-digit Harmonized System (HS6) classification, the broader four-digit (HS4) or two-digit (HS2) categories, or any custom groupings derived from HS6 codes such as green trade or circular economy-associated goods. This specificity bears particular relevance for corporate entities or cluster organizations for whom comprehensive global trade maps may appear overly generalized.

\section{Conclusions}
\label{sec:conclusions}
This paper presented an approach for modeling international trade by leveraging the natural characteristics of Knowledge Graphs to capture contextual and relational information. The integration of the gravity model into the construction process of Knowledge Graphs enables the representation of essential factors influencing trade relationships and facilitates the prediction of future trade patterns.

The research results show how unusual it is to use KG embeddings to predict links and how useful it could be to combine Knowledge Graphs Embeddings with traditional machine learning methods like Decision Trees and Graph Neural Networks. The results indicate improved prediction accuracy and provide a step towards embedding explainability for knowledge representation purposes. The three-dimensional visualization of the generated embeddings offers a valuable tool for understanding and interpreting the underlying trade dynamics.

By employing Knowledge Graphs and Machine Learning techniques, this research contributes to a deeper understanding of international trade flows, supporting decision-making by governments, policymakers, businesses, and researchers. The approach allows for scalable analysis of large and complex trade datasets, facilitating the identification of trends, patterns, and potential impacts of policy changes.

Nevertheless, there are several avenues for future research. Firstly, exploring alternative embedding techniques and incorporating additional domain-specific features could enhance prediction accuracy. Secondly, investigating the interpretability and explainability of KG embeddings in the context of international trade would provide valuable insights and build trust in the decision-making process. In addition, expanding the approach to include more types of data, like transport, economic, and risk indicators, would improve the knowledge representation and give a full picture of how trade works.

The findings of the current research also encourage further interdisciplinary research between economics and data sciences, in particular integrating international trade and product complexity related research streams, adding market demand perspectives, and utilizing explainable artificial intelligence \cite{benyahia2023crossroads, kalvet2023good}.

In conclusion, this research demonstrates the value of knowledge graphs in modeling international trade and highlights the benefits of integrating the gravity model for constructing a robust knowledge base. Combining Knowledge Graphs Embeddings and traditional Machine Learning methods shows promise for improving prediction performance. The findings contribute to the field of knowledge representation and offer practical implications for policymakers and stakeholders in navigating the complex landscape of international trade. By leveraging Knowledge Graphs and machine learning, we can gain deeper insights into trade patterns, anticipate the effects of policy changes, and make more informed decisions to foster economic and social development in the globalized economy.

\section*{Acknowledgements}
This work was supported by grants to TalTech – TalTech Industrial (H2020, grant No 952410) and Estonian Research Council (PRG1573).

 \bibliographystyle{elsarticle-num} 
 \bibliography{bibliography}

\end{document}